\documentclass[lettersize,journal]{IEEEtran}
\usepackage{amsmath,amsfonts}
\usepackage{algorithmic}
\usepackage{algorithm}
\usepackage{array}
\usepackage[caption=false,font=normalsize,labelfont=sf,textfont=sf]{subfig}
\usepackage{textcomp}
\usepackage{stfloats}
\usepackage{url}
\usepackage{verbatim}
\usepackage{graphicx}
\usepackage{cite}
\usepackage{amssymb}
\usepackage{mathtools}
\usepackage{booktabs}
\usepackage{pifont}
\usepackage{float}
\usepackage{bbding}
\usepackage{wrapfig}
\usepackage{multirow}
\usepackage{multicol}
\usepackage[table]{xcolor}
\usepackage{tcolorbox}
\usepackage{hyperref}
\usepackage{mdframed}

\usepackage{algorithm}
\usepackage{booktabs}
\usepackage{amsfonts}
\usepackage{graphicx}
\usepackage{amsmath}
\usepackage{arydshln}
\usepackage{multirow}
\graphicspath{{figures/}}

\usepackage[table]{xcolor}
\usepackage{color}
\usepackage{tcolorbox}
\usepackage{colortbl}

\definecolor{mygray}{gray}{.9}
\usepackage{tcolorbox}

\usepackage{bbding}

\newcommand{\e}[1]{\ensuremath{\times 10^{#1}}}
\newcommand{\ebf}[1]{\ensuremath{\times \mathbf{10^{#1}}}}
\begin{document}

\title{Multi-scale Unified Network for Image Classification}

\author{Wenzhuo Liu, Fei Zhu, Cheng-Lin Liu~\IEEEmembership{Fellow,~IEEE}
\IEEEcompsocitemizethanks{\IEEEcompsocthanksitem The authors are with the School of Artificial Intelligence, University of Chinese Academy of Sciences, Beijing 100049, P.R. China, and the State Key Laboratory of Multimodal Artificial Intelligence Systems, Institute of Automation of Chinese Academy of Sciences, 95 Zhongguancun East Road, Beijing 100190, P.R. China. }
}

\markboth{Journal of \LaTeX\ Class Files,~Vol.~14, No.~8, August~2021}%
{Shell \MakeLowercase{\textit{et al.}}: A Sample Article Using IEEEtran.cls for IEEE Journals}


%

\IEEEtitleabstractindextext{

\begin{abstract}
Convolutional Neural Networks (CNNs) have advanced significantly in visual representation learning and recognition. However, they face notable challenges in performance and computational efficiency when dealing with real-world, multi-scale image inputs.
Conventional methods rescale all input images into a fixed size, wherein a larger fixed size favors performance but rescaling small size images to a larger size incurs digitization noise and increased computation cost.
In this work, we carry out a comprehensive, layer-wise investigation of CNN models in response to scale variation, based on Centered Kernel Alignment (CKA) analysis.
The observations reveal that lower layers are more sensitive to input image scale variations than high-level layers.
Inspired by this insight, we propose Multi-scale Unified Network (MUSN) consisting of multi-scale subnets, a unified network, and scale-invariant constraint. Our method divides the shallow layers into multi-scale subnets to enable feature extraction from multi-scale inputs, and the low-level features are unified in deep layers for extracting high-level semantic features. A scale-invariant constraint is posed to maintain feature consistency across different scales.
Extensive experiments on ImageNet and other scale-diverse datasets,  demonstrate that MSUN achieves significant improvements in both model performance and computational efficiency. Particularly, MSUN yields an accuracy increase up to 44.53\% and diminishes FLOPs by 7.01-16.13\% in multi-scale scenarios.
\end{abstract}

}

\maketitle
\IEEEdisplaynontitleabstractindextext

%
\IEEEpeerreviewmaketitle

\ifCLASSOPTIONcompsoc
\IEEEraisesectionheading{\section{Introduction}\label{sec:introduction}}
\else
\section{Introduction}
\label{sec:introduction}
\fi


\IEEEPARstart
{C}onvolutional neural networks (CNNs) have become the dominant approach for visual representation learning, achieving state-of-the-art performance in various tasks \cite{9474949,7327182,vidalmata2020bridging}. CNNs have proven highly effective in capturing hierarchical features and learning complex patterns in visual data through training on large datasets \cite{9330771,8781933,9495836}. However, despite their remarkable success, CNNs still face performance and computational efficiency challenges when handling inputs of different scales \cite{9650759,8779586,9461631,9052469}.

For convenience of processing by neural networks, the input image is often scaled to a uniform size, such as the widely used \(224 \times 224\) size for ImageNet \cite{krizhevsky2012imagenet}. This practice can be traced back to the pioneering AlexNet, which utilized a consistent input size in the ImageNet classification competition, followed by subsequent architectures like VGGNet \cite{simonyan2014very} and ResNet \cite{he2016deep}. Resizing input images is helpful for improving training efficiency, such as mini-batch learning through gradient descent, which necessitates the same input size for all images in a batch for efficient parallel training of CNNs\cite{alsallakh2023mind}.


However, uniform input size conflicts with real-world scenarios, where images are naturally in variable scales\footnote{The term {\em scale} has many different meanings according to the different factors in imaging. In this paper, we refer {\em scale} specially to the size of image input to neural networks.}, due to the variations of camera devices/parameters, object scale and imaging distance. This discrepancy can severely impact a trained CNN's performance and computational efficiency when applied to different image sizes \cite{9495232,9442912,6338939}. For instance, in our experiments, when testing a ResNet50 model trained on ImageNet resizing images from \(224 \times 224\) to \(32 \times 32\), the accuracy drops from 75.18\% to 19.64\%. Moreover, to adapt to the model's input size, upsampling a \(32 \times 32\) image to \(224 \times 224\) would increase the computational cost by 250.54\%.

\begin{figure}[t]
  \centering
  \vskip -0.1in
  \includegraphics[width=1.0\linewidth]{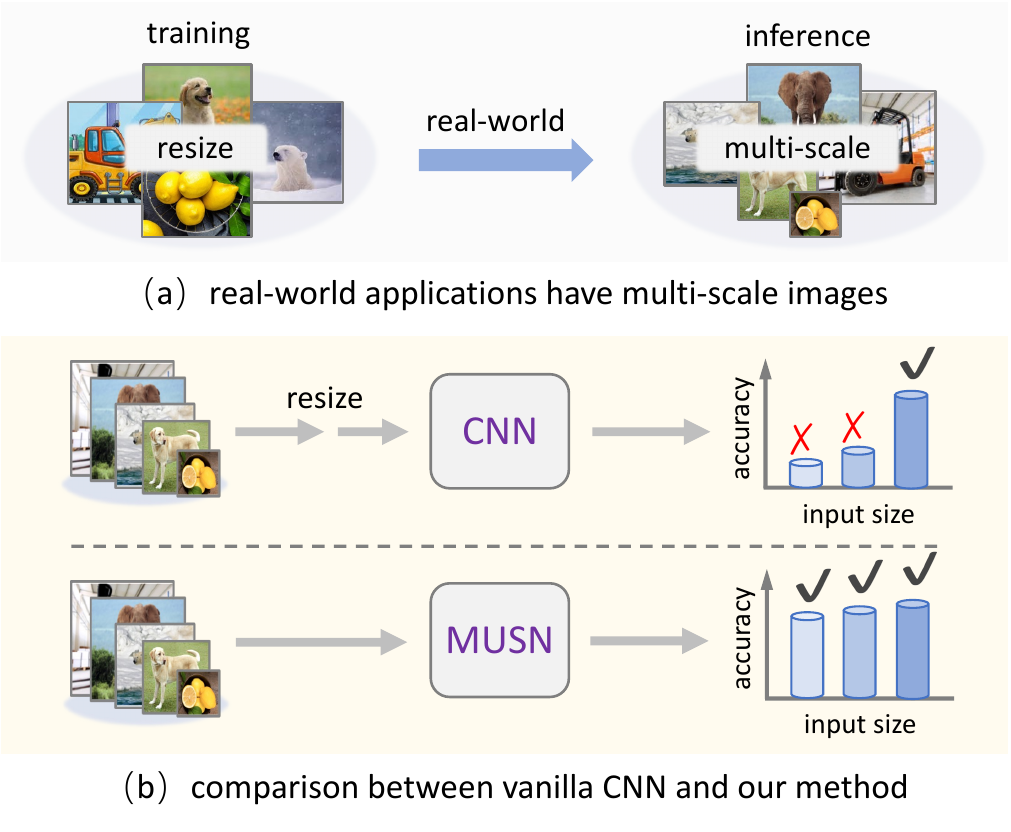}
  \vskip -0.1in
  \caption{In the real world, images come in different sizes, but Vanilla models are designed for a fixed input size, causing performance degradation when the input image scale changes. In contrast, MSUN retains stable performance over a range of scales.}
  \vskip -0.2in
  \label{fig:1}
\end{figure}

A layer-by-layer analysis of CNN models is crucial to address the challenges of performance degradation and computational complexity when handling multi-scale input images. LeCun et al. \cite{lecun2015deep} and Goodfellow et al. \cite{goodfellow2016deep} have provided comprehensive overviews of hierarchical feature extraction capabilities.
It is widely recognized that shallow layers capture low-level features such as edges and textures, while deeper layers extract complex, high-level semantic information. In this work, we analyze the layerwise impacts of CNN models on multi-scale inputs, using Centered Kernel Alignment (CKA) metric \cite{cortes2012algorithms,kornblith2019similarity} to evaluate feature similarity across layers. In analyzing many CNN models like ResNet, DenseNet, VGGNet, and MobileNet, we observe a common phenomenon that \textbf{\textit{lower layers exhibit more pronounced sensitivity to scale variations}}. As image scales change, the substantial alterations in features extracted by these lower layers result in degraded accuracy.

These insights lead us to redesign CNNs to enhance the adaptability of models under multi-scale input images. The primary idea is to separate shallow layers for extracting features at different scales, as they are more sensitive to scale changes. Specifically, distinct lower layers are used to extract features from inputs of different scales, so as to ameliorate the sensitivity to input scale. We then implement scale-invariant constraint to maintain consistency in the features across different scales, an approach commonly applied in traditional handcrafted features such as SIFT \cite{lowe2004distinctive} and SURF \cite{bay2006surf}. After that, a unified network is used to learn high-level semantic information in the higher layers. We refer to this model as Multi-scale Unified Network (MSUN).

Our method can be directly applied to various CNN architectures, significantly enhancing the adaptability in multi-scale scenarios without complicated modifications and increased computational costs. It improves the recognition performance in numerous experiments, including evaluations on ImageNet with varying scales and transfer learning on CIFAR-100\cite{krizhevsky2009learning}, STL-10\cite{coates2011analysis}, Stanford Cars \cite{krause20133d}, and other datasets of different image sizes. The main contributions of this work are summarized as follows:

\begin{itemize}
\item We conduct a layer-wise analysis of CNN's response to input image scale variations. Our investigation reveals that the lower layers of neural networks are highly sensitive to scale changes. This inspires CNN models employing distinct lower layers to extract features for different input scales.
\item Based on this finding, we propose Multi-scale Unified Network to enhance the adaptability of CNNs under varying input scales. It comprises multi-scale subnetworks, a unified network, and scale-invariant constraint. Our method can be readily applied to different CNN architectures.
\item Our method demonstrates improved performance and computational efficiency in extensive experiments, including multi-scale testing on ImageNet and transfer learning on other datasets of different image sizes, thus enhancing the practicality of CNN models in real-world applications.
\end{itemize}

The remainder of this paper is organized as follows. Section \ref{sec: 2} provides an overview of the related work; Section \ref{sec: 3} defines the problem; Section \ref{sec: 4} investigates the layerwise impacts of CNN models on multi-scale input; Section \ref{sec: 5} describes the proposed MSUN in detail. Section \ref{sec: 6} presents our experimental results, and Section \ref{sec: 7} offers concluding remarks.

\section{Related Work}
\label{sec: 2}
\subsection{ Layerwise analysis of CNNs}

CNNs have achieved remarkable results in various visual tasks, but understanding and interpreting CNN models still present some challenges. To address this, researchers have conducted in-depth, layer-by-layer analyses of CNNs, attempting to understand the behavior of these networks by decoupling the features of different layers. Among these studies, a basic consensus is that the shallow layers of deep learning models capture low-level features such as edges and textures, while the deeper layers capture complex, high-level semantic information\cite{lecun2015deep, goodfellow2016deep,bengio2013representation,8116648}. This principle is consistent across a variety of neural networks, including CNNs\cite{8700608,9076866,6522407,7940028}, recurrent neural networks (RNNs)\cite{sutskever2014sequence, cho2014learning, chung2014empirical}, and transformers \cite{vaswani2017attention, radford2018improving, devlin2018bert}.

Some investigations into the transferability of features across CNN layers revealed interesting results.  It was found that initial layers tend to learn more generic features, while the deeper layers become specialized in extracting task-specific characteristics \cite{yosinski2014transferable}. This concept of hierarchical feature learning has inspired the transfer of features learned by CNNs to disparate tasks \cite{donahue2014decaf}. Further, it has been established that a positive correlation exists between the network's depth and its capacity for feature extraction. This underlines the notable proficiency of deeper CNNs in learning intricate representations \cite{szegedy2015going}.

In addition to the fundamental functionality of CNNs, many methods have been proposed to visualize and understand their internal dynamics. For example, the deconvolutional method \cite{zeiler2014visualizing} was used to scrutinize the different layers of a CNN. The Deep Inside Convolutional Networks (DICN)  \cite{simonyan2013deep} offered an in-depth view of CNNs' intermediate activations. The introduction of Network Dissection \cite{bau2017network} allowed for quantitative interpretation of individual CNN units, revealing the upper layers' ability to represent object components or even entire objects. Furthermore, guided backpropagation \cite{springenberg2014striving} was employed to highlight class-discriminative regions in CNNs, illuminating the distinct roles of different layers.

However, a layer-wise examination of network behavior under scale variation is unexplored, despite its practical importance for addressing the challenges of varying input scales in CNNs. In this work, we use the Centered Kernel Alignment (CKA)  to quantitatively analyze different layers under scale changes. Our experiment reveals that in many representative CNN models, the lower layer features exhibit a more pronounced impact from scale changes, thus leading to degraded network performance under scale variations. This insight inspires our method of training different lower layers of the network for multi-scale input.

\subsection{Scale-invariant Feature Extraction}

Scale-invariant feature extraction aims to extract features consistent across variable image scales, enhancing the model's robustness to changes in image scale. This concept has been integral to traditional features since the early research, methods such as Scale-Invariant Feature Transform (SIFT) \cite{lowe2004distinctive} and Speeded Up Robust Features (SURF) \cite{bay2006surf}. SIFT achieves scale invariance by processing the image in a scale space generated through Gaussian blurring and downsampling of the original image. During this process, SIFT identifies extrema points in the scale space, which serve as keypoints. SURF utilizes the Hessian matrix for keypoint detection and employs the integral image to compute the Hessian matrix at different scales, significantly improving computational efficiency. As a result, both SIFT and SURF are capable of locating these keypoints regardless of image scale.

After that, Affine-Scale Invariant Feature Transform (ASIFT) extends the scale invariance of SIFT by introducing affine invariance \cite{morel2009asift}. Maximally Stable Extremal Regions (MSER) provides scale invariance through region-based feature representation \cite{matas2004robust}. Local Binary Descriptors (LDB) and Binary Robust Invariant Scalable Keypoints (BRISK) employ binary descriptors to improve speed and robustness \cite{yang2013local, leutenegger2011brisk}. These works continuously improve and explore scale invariance in feature extraction, enhancing matching performance and computational efficiency.

Traditional methods have inspired deep learning models to incorporate scale invariance in their design. For instance, Scale-invariant CNN (SiCNN) \cite{xu2014scale} encodes scale information into the architecture through shared convolutional kernels across scales. The Inception module \cite{szegedy2015going}, which uses parallel convolutional layers of varying kernel sizes to capture different scale features, has been adopted widely in popular architectures like GoogLeNet \cite{szegedy2015going} and InceptionV3 \cite{szegedy2016rethinking}. ZoomNet \cite{wang2017zoom} adopts multi-scale feature fusion for learning across scales. The Pyramid Scene Parsing Network (PSPNet) \cite{zhao2017pyramid} employs pyramid pooling to capture global context information at various scales. Feature Pyramid Networks (FPN) \cite{lin2017feature} construct a top-down architecture with lateral connections to blend high-level semantic information with low-level spatial information, forming a multi-scale feature pyramid. Atrous convolutions \cite{chen2017deeplab}, by adjusting the dilation rate, control the effective field of view, capturing context information at varying scales without parameter increase. This technique has seen successful application in the DeepLab architectures \cite{chen2017deeplab, chen2018encoder}. 

The above models reveal two shortcomings, however. \textbf{Firstly}, their performance tends to significantly decrease when the scale of the input images changes. They are not aimed to extract unified features robust to input scale variation, but to provide multi-hierarchy features for handling different scales of objects. This is different from the scale invariance we originally hoped to achieve. \textbf{Secondly}, different layers of neural networks have unique functions and characteristics, and these should be considered cooperatively with scale invariance constraints. Our experiments show that the lower layers of the network are more sensitive to scale changes, and they should be prioritized when satisfying scale invariance.

\subsection{Multi-Scale Adaptive CNNs}

Supporting multi-resolution input in neural networks has been an issue of active research, as it offers potential improvements in computational efficiency and robustness for neural networks. A direct idea is to train neural networks using images of different resolutions (still resized to fixed-size input, however). which is called multi-scale data augmentations or multi-scale training.  This method encourages models to learn features across scales, establishing a widely adopted paradigm\cite{chen2018gradually,zhao2018multi, chen2018gradually, zhao2018multi}. However, training on mixed-scale images can be detrimental to the performance on standard resolutions.  Another method is to learn how to resize images\cite{cohen2020learning}. It can notably improve model performance at a specific resolution by replacing the conventional bilinear interpolation with a learned, CNN-based resizer integrated into the training process. Despite this, the trained resizer model can only adjust images to a certain resolution, exhibiting limited adaptability across different scales.

Some methods aim to balance between computational efficiency and accuracy by altering image resolution and network structure, such as Resolution Adaptive Networks (RANet)\cite{wang2018resolution}, Multi-Scale Dense Networks (MSDNet)\cite{huang2018multi}, and Dynamic Resolution Networks (DRNet)\cite{chen2020dynamic}, adapt the model's architecture and resolution dynamically based on whether the input image is easy or difficult to infer. Among them, RANet downsamples easy-to-judge samples and uses a smaller network for inference, thereby improving its efficiency. DRNet uses a Resolution Predictor to predict the image resolution to be downscaled based on difficulty. MSDNet optimizes computational efficiency on simpler images by downsampling them and predicting based on the output of the network's mid-layers, rather than using the entire network. These methods improve the computation efficiency by adaptively downsampling easy images, but do not consider the scale invariance of input images which are originally multi-scaled or multi-resolution.

In this work, we propose a simple but effective method that can be applied to existing CNNs without complicated modifications. Distinct from previous methods, it considers inherent scale invariance of multi-scale input images in the design of network architecture and training algorithm. The proposed MSUN achieves remarkable improvements in computational efficiency and overall performance in experiments on multiple datasets of different image sizes.

\section{Preliminaries}
In real-world applications, images input to deep learning models often have variable scales. This necessitates that models maintain robust performance across diverse input scales without retraining.  Conventional methods re-size all input images into the same size for convenient processing by neural networks. This may deteriorate the performance on images of variable scales, however.

To assess the model performance across different scales of input, let's consider a standard classification task involving a training dataset $D_{\text{train}}$ and a testing dataset $D_{\text{test}}$.
In the multi-scale setup, we resize $D$ into a collection of testing sets with varying scale: $D_0, D_1,..., D_N$,corresponding to input sizes $R_0, R_1, \dots, R_N$. The model is then evaluated on these testing sets to gauge its performance under each size, instead of just evaluating it at the image size consistent with the training set.

We consider two settings for multi-scale experiments:
\begin{itemize}
\item \textbf{Multi-scale testing:} In this case, the categories $C$ of $D_{\text{train}}$ and $D_{\text{test}}$ remain the same. Compared to the standard classification task, the model's performance is evaluated on a broader image size range of $D_{\text{test}}$, in different input sizes $R_0, R_1, \dots, R_N$.
\item \textbf{Multi-scale transfer learning:} For a model trained on $D_{\text{train}}$, $D_{\text{test}}$ possesses different categories and distributions from $D_{\text{train}}$. This evaluates the transferability of the trained model on datasets of different sizes.
\end{itemize}
We represent the model with two parts for analysis: a feature extractor $\mathcal{F}$ and a classifier $g$. The feature extractor, parameterized by $\theta$, encodes the input into a representation as $z = \mathcal{F}(x; \theta)$. In multi-scale testing, the trained $\mathcal{F}$ and $g$ are directly used for evaluation. In multi-scale transfer learning, we train the feature extractor and classifier on the transfer dataset via fine-tuning. For linear evaluation, only $g$ is trained while $\mathcal{F}$ is kept fixed.

\label{sec: 3}

\begin{figure*}[t]
  \centering
  \includegraphics[width=1.0\linewidth]{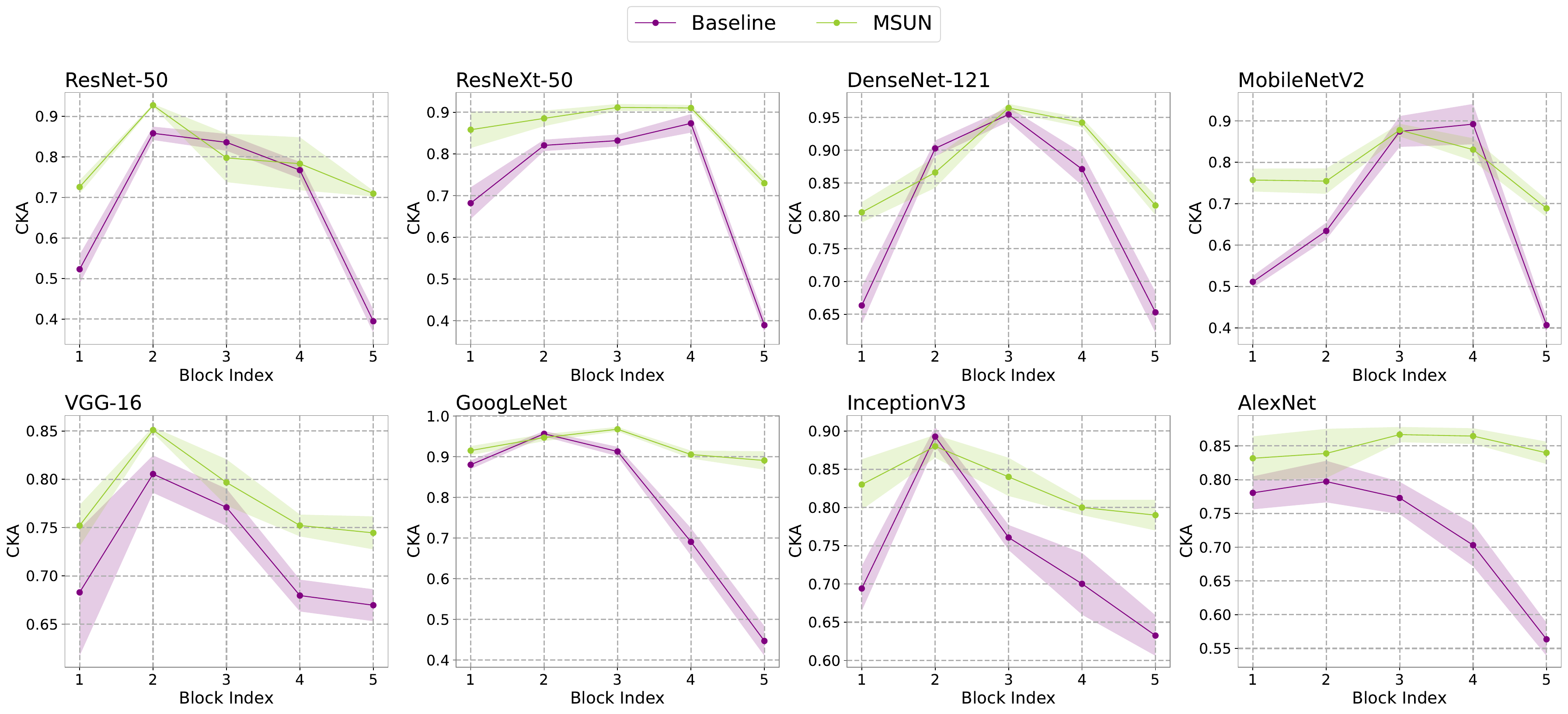}
  \vskip -0.1in
\caption{The layer-wise feature similarity between \(32 \times 32\) and \(224 \times 224\) inputs, showing that the lower layers in these CNN models are considerably more sensitive to scale changes. Compared to the baseline, our method (MSUN)  maintains a higher CKA between inputs of different input scales, obtaining more robust representations across scale variations.}
  \vskip -0.13in
  \label{fig:2}
\end{figure*}

\section{Layer-wised Investigation of CNNs under Multi-scale Input}
\label{sec: 4}
\subsection{Performance Collapse of Scale Variations}
Current CNNs maintain uniform image size within a batch during training to take advantage of parallel processing efficiencies. However, in the real world, images present a wide range of sizes,  and when a well-trained network is deployed, it is expected to grapple with inputs of varying scales.

Intuitively, there are two strategies for tackling this: apply the CNN directly to the original image size (this needs resizing the training images to match the size of test images), or resize the original image to match the model's input size of training data before inference.
We conduct experiments on a range of commonly-used networks, including ResNet \cite{he2016deep}, ResNeXt \cite{xie2017aggregated}, DenseNet \cite{huang2017densely}, MobileNetV2 \cite{sandler2018mobilenetv2}, VGGNet \cite{simonyan2014very}, GoogleNet \cite{szegedy2015going}, InceptionV3 \cite{szegedy2016rethinking}, and AlexNet \cite{krizhevsky2012imagenet}.
To simulate and assess how these models perform when handling varying scale images in real world, the ImageNet test set is resized to different image size, like \(32 \times 32\), \(128 \times 128\) and \(224 \times 224\).
The results of these two strategies shown in Table \ref{table:1} highlight two key findings:

\textbf{First,} models inputting images of lower image size result in notable performance drop. This issue exists even in GoogleNet and InceptionV3, which aim for scale invariance. Lower image size leads to low performance because the image loses many object details in low resolution.

\textbf{Second,} compared to inference at the original image size, resizing images to match the training set image size boosts performance, improving accuracy by 3.09\%-12.70\% and 10.50\%-21.71\%. However, this increases the computational overhead of the model.

As shown in Table \ref{table:1}, when the input size is reduced from \(224 \times 224\) to \(128 \times 128\), the model performance drops by 4.64\%-7.01\%, when the input size is further reduced to \(32 \times 32\), the models'  suffer from performance drop as large as 43.09\%-59.97\%.

\begin{table}[t]

   \caption{Performance evaluation of CNNs on different input sizes of ImageNet dataset. An asterisk (*) signifies inference on the original image size (by resizing training images to the test image size), whereas non-asterisk columns indicate that the images are resized to \(224 \times 224\).Some results are missing for Inception-v3 and AlexNet, which do not work on these reduced image sizes.}
    \label{table:1}
       \vskip -0.1in
 \small \centering 
 \renewcommand\tabcolsep{3.5pt}
 \renewcommand{\arraystretch}{1.2}
\begin{tabular}{lccccc}
\toprule 
\multirow{2}{*}{Models} & \multicolumn{5}{c}{Input size}       \\
                        & 32x32*  & 32x32    & 128x128*  & 128x128   & 224x224   \\ \hline \midrule
ResNet50                & 4.48 & 19.64 & 64.68 & 68.84 & 75.18 \\
ResNext50               & 3.64 & 25.35 & 67.63 & 72.92 & 77.61 \\
DenseNet121             & 3.63 & 21.07 & 63.61 & 69.26 & 74.51 \\
VGG16                   & 3.85 & 14.35 & 64.22 & 67.31 & 74.32 \\
MobileNet-v2            & 1.63 & 16.16 & 58.70 & 65.23 & 71.97 \\
GoogleNet               & 1.62 & 15.47 & 54.91 & 64.52 & 69.77 \\
Inception-v3            & -    & 21.41 & -     & 64.88 & 69.52 \\
AlexNet                 & -    & 13.46 & 37.83 & 50.53 & 56.55 \\ 
  \bottomrule
\end{tabular}
\end{table}
\subsection{Centered Kernel Alignment}
To analyze the effects of different layers on model performance changing input scale, we introduce the Centered Kernel Alignment (CKA) metric \cite{cortes2012algorithms,kornblith2019similarity}, a widely used method to compare the learned representations across different layers of neural networks.  CKA computes the similarity between two sets of features by comparing their centered Gram matrices, which are the inner products of the feature vectors after centering.

Consider two sets of feature representations, $X \in \mathbb{R}^{n \times d_x}$ and $Y \in \mathbb{R}^{n \times d_y}$, where $n$ is the number of samples and $d_x$ and $d_y$ signify the dimensionalities of the features. The first step in the CKA metric computation is to center these feature representations by subtracting their means:
\begin{align*}
\tilde{X} &= X - \frac{1}{n} X \mathbf{1}{n \times n}, \\
\tilde{Y} &= Y - \frac{1}{n} Y \mathbf{1}{n \times n},
\end{align*}
where $\mathbf{1}_{n \times n}$ is an $n \times n$ matrix with all elements equal to $\frac{1}{n}$.
The centered Gram matrices, $K_X$ and $K_Y$, are then calculated as the inner products of the centered feature representations:
\begin{align*}
K_X &= \tilde{X} \tilde{X}^T, \\
K_Y &= \tilde{Y} \tilde{Y}^T.
\end{align*}
Finally, The CKA metric is computed as the normalized Frobenius inner product of the centered Gram matrices:
\begin{equation}
\text{CKA}(X, Y) = \frac{\langle K_X, K_Y \rangle_F}{|K_X|_F |K_Y|_F},
\end{equation}
where $\langle \cdot, \cdot \rangle_F$ denotes the Frobenius inner product, and $|\cdot|_F$ denotes the Frobenius norm.

The value of CKA ranges from 0 to 1, with 0 indicating no similarity between the feature representations and 1 indicating identical feature representations.

\subsection{Understanding the Impact of Scale Variations}

To understand the performance associated degradation with scale variations, we initiate a layer-wise inspection using CKA. In particular, we evaluate a convolutional network $F$ divided into five intermediate layers or blocks, denoted as $F_1, F_2, F_3, F_4,$ and $F_5$. We record the outputs, $F_i(x)$, of these intermediate layers and compare the similarity of these outputs under small (\(32 \times 32\)) and large (\(224 \times 224\)) scale inputs, denoted by $X_{32}$ and $X_{224}$ respectively, using the following equation:
\begin{equation}
CKA(F_i(X_{32}), F_i(X_{224})) \quad \text{for} \quad i = 1,2,...,5.
\end{equation}
The CKA curve offers a quantitative indication of feature variations within the network. Intuitively, a robust model should maintain a high feature similarity under diverse scale inputs. As shown in Figure 2, the curve of a better model should be closer to 1, especially in the final layer. 
We observe the following phenomena of these CNN models:
\begin{itemize}
\item The features of the final layer, which are directly associated with classification performance, show remarkable shifts when dealing with scale variations. This denotes a lack of robustness in these models, leading to the performance corruption shown in Table \ref{table:1}.
\item Across these models, it is commonly observed that \textbf{\textit{the lower layers are more heavily impacted by scale variations compared to other intermediate layers}}. The lower layer features affect the final layer features through forward propagation, thus leading to degradation of final performance
\end{itemize}
This trend is also in line with previous analyses on CNNs \cite{lecun2015deep, goodfellow2016deep,bengio2013representation}. They suggest that lower layers extract local features such as edges and textures, whereas deeper layers capture more global, semantic features. When the input scale changes, the local details of the image are modified, but the global semantics should stay consistent across scales. 
Therefore, performance corruption can be attributed to lower layers' sensitivity to scale changes. This motivates our work to handle scale variation in model design.

Our method employs different subnetworks to handle inputs of different scales. It poses consistency in the lower layer features across scales, to enhance the CKA similarity of both lower and final layers is achieved. As illustrated in Figure \ref{fig:2}, This denotes an overall improvement in robustness to multi-scale inputs.  A detailed description of our method will be provided in Section \ref{sec: 5}.

\begin{figure*}[t]
  \centering
  \includegraphics[width=0.95\linewidth]{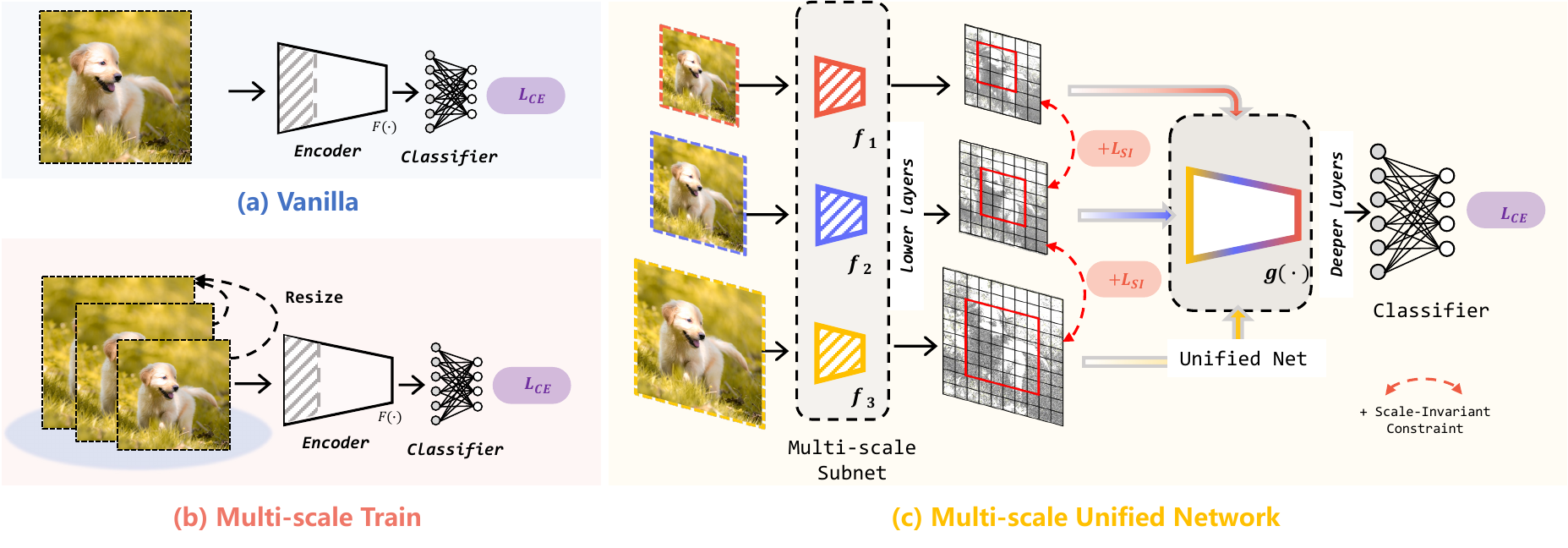}
  \vskip -0.1in
  \caption{Illustration of the Vanilla model\cite{he2016deep,huang2017densely,simonyan2014very,sandler2018mobilenetv2}, multi-scale training, and our method. MSUN has the lower layers divided into multi-scale subnets, which are unified in upper layers incorporating a scale-invariance constraint.}
  \vskip -0.1in
  \label{fig:3}
\end{figure*}

\section{Method: Multi-scale Unified Network}
\label{sec: 5}
\subsection{Overview of the Method}
The proposed Multi-scale Unified Network (MSUN) consists of multi-scale subnetworks, scale-invariant constraints, and a unified network. MSUN is compatible with most widely used CNN architectures, including ResNet, DenseNet, ResNext, VGGNet, MobileNetV2, etc. It is designed to provide a more flexible and robust way of handling varying input scales.

The shallow layers of MSUN are partitioned into multi-scale subnetworks designed to handle different input scales divided into rough categories such as large, medium, and small. The subnetworks independently extract low-level features, to overcome the sensitivity to scale variations. To extract robust high-level features, we enforce scale-invariant constraints to ensure consistency in the features extracted across different input scales, which is crucial in enhancing the network's resilience to changes in scale. The extracted low-level features are then processed through a common deep network to extract high-level semantic features.
In the inference phase, the subnetwork aligning with the input image's scale is selected, avoiding the computational costs of rescaling input image, especially into a larger size.

Our method is illustrated in Figure \ref{fig:3}. It differs from the Vanilla model (which has fixed input size) and multi-scale training method (which transforms multi-scale images into the same size in training and inference) in that it has multiple lower-layer subnetworks and a unified high-level network for enhancing scale invariance and avoiding rescaling input images to a fixed size.

\subsection{Multi-scale Subnet and Unified Net}

To apply MSUN on a standard CNN model $F$, which consists of lower layers $f$ and deeper layers $g$, expressed as $\mathcal{F}(x)=g(f(x))$, parallel subnetworks are used to replace the lower layers $f$, aiming to extract features at varying scales. In this paper, we utilize three quantized scales (large, medium, and small) which turn out to perform sufficiently well. The large-scale subnetwork aligns with the shallow network $f$ in the original model. For the other two scales, we substitute the large convolutions with smaller ones and discard some max-pooling layers, intending to make it adapt to smaller input sizes. Subsequently, the unified network $g$ extracts high-level semantic features from each scale subnetwork's output.
The overall feature extraction process of the ordinary CNN model $\mathcal{F}$ and the MSUN $\widehat{\mathcal{F}}$ are formulated as:

\begin{align}
\mathcal{F}(x) &= g(f_N(x_N)), \\
\widehat{\mathcal{F}}(x) &= [g(f_1(x_1)),g(f_2(x_2)),...,g(f_N(x_N))], 
\end{align}
where $x_i$ represents the $i$-th quantized scale of input image and $f_i$ denotes the subnetwork corresponding to the $i$-th scale. MSUN assumes $N$ scales ($N=3$ in our paper), while the ordinary CNN directly takes the largest scale $x_N$ for high performance.

Given an input $x$ with a certain size $R(x)$ in the inference stage, the model resizes the image to the quantized scale $R(x_i)$ that closely matches a subnetwork. The model only needs to calculate the low-level features at this scale, which aligns with the computation process of the original CNN model: 
\begin{align}
\widehat{\mathcal{F}}_{\text{test}}(x) &= g( f_i(x_i) )\\
i&=\arg\min_{i=1}^N(|R(x)-R(x_i)|).
\end{align}
The architecture of multi-scale subnetworks and a unified net provides a more flexible approach for handling varying input scales. This approach enables calculations to be performed at a size much closer to that of the original image, significantly improving computational efficiency when dealing with multi-scale inputs. Consequently, this flexibility also improves model performance due to the better adaptability to input scale variation.

\begin{algorithm}[t]
\label{algorithm:0}
\caption{Multi-Scale Unified Network Training}
\begin{algorithmic}[1]
\STATE \textbf{Input:} CNN model $\mathcal{F}$, training images $x$, labels $y$
\STATE Transform F to subnetworks $f_1, f_2, \dots, f_n$ and unified network $g$
\FOR {each epoch}
    \FOR{each mini-batch}
        \STATE Calculate $L_{CE}$ for each scale $x_i$ by forward-passing the input through $f_i$ and $g$
        \STATE Calculate $L_{SI}$ by computing $D(f_i(x), f_j(x_s))$ for all pairs of scales
        \STATE Calculate overall loss $L = L_{CE} + \max(\lambda, L_{SI})$
        \STATE Backpropagate the gradients and update the parameters of $f_i$ and $g$
    \ENDFOR
\ENDFOR
\end{algorithmic}
\end{algorithm}

\begin{figure*}[htb]
  \centering
  \includegraphics[width=1.0\linewidth]{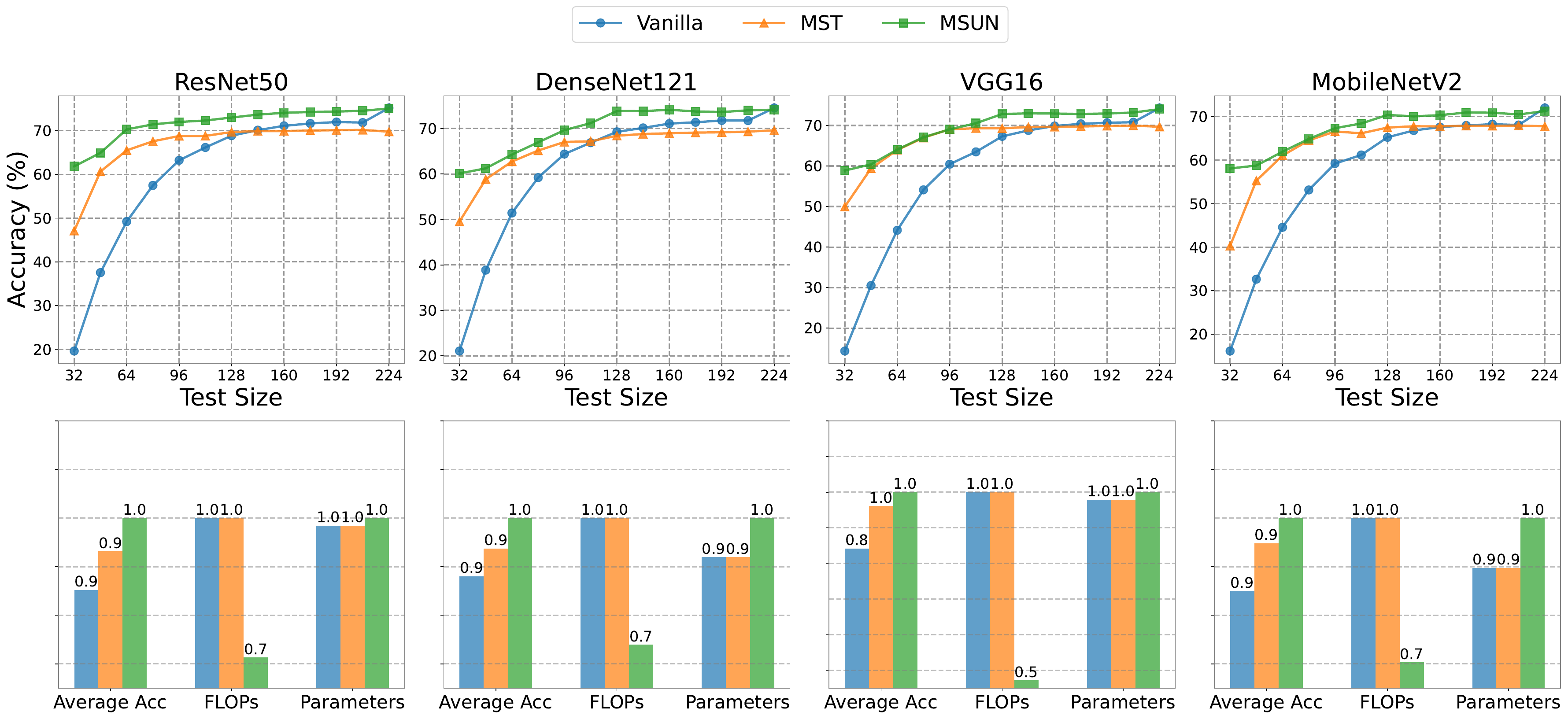}
  \vskip -0.1in
  \caption{Multi-scale testing on ImageNet spanning from \(32 \times 32\) to \(224 \times 224\) image size with a stride of 16. (a) The model's accuracy curves at each input size. (b) The model's average accuracy, average FLOPs, and total model parameters across these sizes. Our method demonstrates significant accuracy improvements and reduced FLOPs, mitigating the model's performance breakdown at lower image scales.}
  \vskip -0.13in
  \label{fig:4}
\end{figure*}
\subsection{Scale-Invariant Constraint}
To make features extracted by the multi-scale subnetworks consistent across scale space, we introduce a scale-invariant constraint inspired by traditional features extensively used in the past, such as SIFT \cite{lowe2004distinctive} and SURF \cite{bay2006surf}, through the creation of a scale space and assuming invariance across difference scales.

For achieving scale invariance in the unified high-level layers of MSUN, we introduce the scale-invariant constraint (SI) as a regularization term, $L_{SI}(f_1, f_2, \dots, f_n)$, in the loss function. In this context, $x_i$ represents an image $x$ resized to the $i$-th quantized scale, while $f_i$ denotes the shallow subnetwork corresponding to the $i$-th scale.
The SI is calculated as:
\begin{equation}
L_{SI}(f_1, f_2, \dots, f_n) = \sum_{i=1}^{n-1} \sum_{j=i+1}^{n} D(f_i(x), f_j(x_s)),
\end{equation}
where $D(\cdot, \cdot)$ represents a distance metric (e.g., Euclidean distance) that measures the discrepancy between the features extracted by different subnetworks.

The overall loss function for training MSUN models is:
\begin{equation}
L = \max(L_{SI}(f_1, f_2, \dots, f_n),\lambda)+\sum_{i=1}^NL_{CE}(g(f_i(x)), y),
\end{equation}
where $L_{CE}$ denotes the cross-entropy loss function for classification, $y$ represents the ground truth labels of $x$, and $\lambda$ is a hyperparameter for balancing task-specific loss and the scale-invariant constraint.

Algorithm \ref{algorithm:0} provides the pseudo-code for our method. By incorporating the SI constraint into the loss function, the CNN model is encouraged to learn features robust to scale variation, leading to improved performance when dealing with images of varying scales during inference.

\section{Experiments}
\label{sec: 6}
\subsection{Evaluation Metrics}
The performance metrics for evaluating models under multi-scale inputs are based on the Learning to Resize method \cite{cohen2020learning} and other related works \cite{alsallakh2023mind,touvron2019fixing}. We use three main metrics to assess our method:

\noindent\textbf{Average Accuracy:} This metric calculates the mean of accuracies achieved across all test input sizes: $\bar{A} = \frac{1}{T}\sum_{t=1}^{T}{A}_t$, where $A_t$ is the classification accuracy evaluated on the current test set at the $t$-th scale.

\noindent\textbf{Parameters:} The model complexity is measured by the number of trainable parameters in the model.

\noindent\textbf{FLOPs:} The metric Floating Point Operations Per Second (FLOPs) is used to evaluate the computational cost of the model. FLOPs of each layer is defined as $FLOPs = 2 \times n \times m \times k^2$, where $n$ denotes the number of input feature maps, $m$ denotes the number of output feature maps, and $k$ represents the size of the convolutional kernel.

\subsection{Setup}

\noindent\textbf{Datasets:} Our experiments used diverse datasets, with ImageNet \cite{deng2009imagenet} for primary training. Validation images were resized to \(32 \times 32\) to \(224 \times 224\) (step 16) pixels in multi-scale testing. The model trained on ImageNet was transferred to datasets CIFAR-10 \cite{krizhevsky2009learning}, CIFAR-100 \cite{krizhevsky2009learning}, STL-10 \cite{coates2011analysis}, Caltech-101 \cite{fei2007learning}, Fashion-MNIST \cite{xiao2017fashion}, Oxford 102 Flowers \cite{nilsback2008automated}, Oxford-IIIT Pets \cite{parkhi2012cats}, Stanford Cars \cite{krause20133d}, FGVC Aircraft\cite{maji2013fine}, Describable Textures Dataset (DTD) \cite{cimpoi2014describing}, and RAF-DB \cite{li2018reliable}, spanning image sizes of \(28 \times 28\) to \(224 \times 224\) pixels. We used the first train/test split for DTD, and random 30 images per class for Caltech-101 testing.

\noindent\textbf{Backbone Networks:} Our method is validated on popular CNN architectures including ResNet-50 \cite{he2016deep}, DenseNet-121 \cite{huang2017densely}, VGG-16 \cite{simonyan2014very}, and MobileNet-v2 \cite{sandler2018mobilenetv2}. We compare MSUN with the original models (Vanilla), and models with multi-scale training (MST). These CNNs are utilized in multi-scale testing and transfer learning. Ablation studies are conducted using ResNet-50.

\noindent\textbf{Implementation Details:} Networks are trained using the Stochastic Gradient Descent (SGD) optimizer \cite{krizhevsky2012imagenet} with an initial learning rate of 0.1, momentum of 0.9, weight decay of 2e-5, and a batch size of 128 per GPU. The learning rate scheduler employed is LinearWarmupCosineAnnealingLR with a 5-epoch warm-up and start/minimum learning rates at 0.01 of the initial rate.
In our network architecture, we employ three quantized scales: large (\(224 \times 224\)), medium (\(128 \times 128\)), and small (\(32 \times 32\)).
Networks are trained for 90 epochs, with MobileNetV2 extended to 300 epochs. Linear-probe and fine-tuning run for 30 epochs. Other settings like data transformation follow PyTorch's official implementation. \textit{Our code will be made public once this paper is accepted, and more implementation details can be found in the code.}

\begin{table*}[t]
\caption{Multi-scale testing on ImageNet across input sizes from \(32 \times 32\) to \(224 \times 224\) with a stride of 16. The measures assessed include the accuracy at each size, average FLOPs, and the number of model parameters.}
\label{table:2}
\small
 \centering 
 \renewcommand\tabcolsep{2.5pt}
 \renewcommand{\arraystretch}{1.2}	
\begin{tabular}{lcccccccccccccccc}
\toprule 
\multicolumn{2}{c}{}                                        &                                &                                           & \multicolumn{13}{c}{\textbf{Input size}}                                                                                                                                                                                                                                                                                                                                                                                                                                                                                                           \\
\cline{5-17}
\multicolumn{2}{c}{\multirow{-2}{*}{\textbf{Methods}}}               & \multirow{-2}{*}{\textbf{Params ($\downarrow$)}}       & \multirow{-2}{*}{\textbf{FLOPs ($\downarrow$)}}                   & 32                                     & 48                                     & 64                                     & 80                                     & 96                                     & 112                                    & 128                                    & 144                                    & 160                                    & 176                                    & 192                                    & 208                                    & 224                           \\ \hline\midrule
                              & Vanilla                     & \textbf{25.6M}                 & 3.28\e{10}                                  & 19.64                                  & 37.55                                  & 49.22                                  & 57.48                                  & 63.22                                  & 66.15                                  & 68.84                                  & 70.11                                  & 71.09                                  & 71.64                                  & 71.96                                  & 71.84                                  & \textbf{75.18}                \\
                              & MST                         & \textbf{25.6M}                          & 3.28\e{10}                                  & 47.11                                  & 60.65                                  & 65.49                                  & 67.56                                  & 68.77                                  & 68.80                                  & 69.67                                  & 69.89                                  & 69.90                                  & 70.02                                  & 70.10                                  & 70.15                                  & 69.76                         \\
\multirow{-3}{*}{ResNet50}    & \cellcolor{mygray}MSUN & \cellcolor{mygray}26.0M  & \cellcolor{mygray}\textbf{3.05\ebf{10}} & \cellcolor{mygray}\textbf{61.83} & \cellcolor{mygray}\textbf{64.92} & \cellcolor{mygray}\textbf{70.28} & \cellcolor{mygray}\textbf{71.43} & \cellcolor{mygray}\textbf{71.98} & \cellcolor{mygray}\textbf{72.31} & \cellcolor{mygray}\textbf{72.99} & \cellcolor{mygray}\textbf{73.66} & \cellcolor{mygray}\textbf{74.07} & \cellcolor{mygray}\textbf{74.26} & \cellcolor{mygray}\textbf{74.35} & \cellcolor{mygray}\textbf{74.53} & \cellcolor{mygray}75.06 \\ 
\hline
                              & Vanilla                     & \textbf{8.0M}                  & 2.28\e{10}                                  & 21.07                                  & 38.86                                  & 51.42                                  & 59.20                                  & 64.38                                  & 66.87                                  & 69.26                                  & 70.12                                  & 71.07                                  & 71.36                                  & 71.76                                  & 71.73                                  & \textbf{74.51}                \\
                              & MST                         & \textbf{8.0M}                           & 2.28\e{10}                                  & 49.53                                  & 58.83                                  & 62.73                                  & 65.15                                  & 67.03                                  & 67.18                                  & 68.40                                  & 68.78                                  & 68.94                                  & 69.09                                  & 69.16                                  & 69.31                                  & 69.59                         \\
\multirow{-3}{*}{\begin{tabular}[c]{@{}c@{}}DenseNet\\ 121\end{tabular}} & \cellcolor{mygray}MSUN & \cellcolor{mygray}8.7M   & \cellcolor{mygray}\textbf{1.99\ebf{10}} & \cellcolor{mygray}\textbf{60.11} & \cellcolor{mygray}\textbf{61.23} & \cellcolor{mygray}\textbf{64.27} & \cellcolor{mygray}\textbf{66.95} & \cellcolor{mygray}\textbf{69.63} & \cellcolor{mygray}\textbf{71.17} & \cellcolor{mygray}\textbf{73.83} & \cellcolor{mygray}\textbf{73.81} & \cellcolor{mygray}\textbf{74.11} & \cellcolor{mygray}\textbf{73.74} & \cellcolor{mygray}\textbf{73.61} & \cellcolor{mygray}\textbf{73.99} & \cellcolor{mygray}74.12 \\ 
\hline
                              & Vanilla                     & \textbf{138.0M}                & 1.24\e{11}                                  & 14.35                                  & 30.51                                  & 44.15                                  & 54.11                                  & 60.45                                  & 63.48                                  & 67.31                                  & 68.78                                  & 69.87                                  & 70.40                                  & 70.67                                  & 70.80                                  & \textbf{74.32}                \\
                              & MST                         & \textbf{138.0M}                         & 1.24\e{11}                                  & 49.96                                  & 59.41                                  & 63.96                                  & 66.97                                  & 69.07                                  & 69.31                                  & 69.30                                  & 69.60                                  & 69.64                                  & 69.75                                  & 69.88                                  & 69.97                                  & 69.68                         \\
\multirow{-3}{*}{VGG16}       & \cellcolor{mygray}MSUN & \cellcolor{mygray}141.0M & \cellcolor{mygray}\textbf{1.04\ebf{11}} & \cellcolor{mygray}\textbf{58.88} & \cellcolor{mygray}\textbf{60.40} & \cellcolor{mygray}\textbf{64.05} & \cellcolor{mygray}\textbf{67.10} & \cellcolor{mygray}\textbf{69.04} & \cellcolor{mygray}\textbf{70.55} & \cellcolor{mygray}\textbf{72.83} & \cellcolor{mygray}\textbf{72.98} & \cellcolor{mygray}\textbf{72.94} & \cellcolor{mygray}\textbf{72.83} & \cellcolor{mygray}\textbf{72.96} & \cellcolor{mygray}\textbf{73.19} & \cellcolor{mygray}74.03 \\ 
\hline
                              & Vanilla                     & \textbf{3.5M}                  & 2.46\e{9}                                  & 16.16                                  & 32.65                                  & 44.57                                  & 53.14                                  & 59.20                                  & 61.17                                  & 65.23                                  & 66.80                                  & 67.58                                  & 67.95                                  & 68.24                                  & 68.05                                  & \textbf{71.97}                \\
                              & MST                         & \textbf{3.5M}                           & 2.46\e{9}                                  & 40.29                                  & 55.27                                  & 61.04                                  & 64.53                                  & 66.57                                  & 66.15                                  & 67.47                                  & 67.76                                  & 67.75                                  & 67.88                                  & 67.86                                  & 67.94                                  & 67.75                         \\
\multirow{-3}{*}{\begin{tabular}[c]{@{}c@{}}MobileNet\\ V2\end{tabular}} & \cellcolor{mygray}MSUN & \cellcolor{mygray}3.9M   & \cellcolor{mygray}\textbf{2.28\ebf{9}} & \cellcolor{mygray}\textbf{58.07} & \cellcolor{mygray}\textbf{58.74} & \cellcolor{mygray}\textbf{61.93} & \cellcolor{mygray}\textbf{64.85} & \cellcolor{mygray}\textbf{67.33} & \cellcolor{mygray}\textbf{68.40} & \cellcolor{mygray}\textbf{70.37} & \cellcolor{mygray}\textbf{70.04} & \cellcolor{mygray}\textbf{70.31} & \cellcolor{mygray}\textbf{70.93} & \cellcolor{mygray}\textbf{70.88} & \cellcolor{mygray}\textbf{70.48} & \cellcolor{mygray}71.24 \\ 
\bottomrule 
\end{tabular}
 \vskip -0.15in
\end{table*}

\begin{table*}[htb]
\caption{Comparison of transfer learning performance using our method, Vanilla baselines, and multi-scale training, evaluated over 11 natural image datasets. The input size varies from 28x28 to \(224 \times 224\), deploying models pre-trained on ImageNet.}
 \label{table:4}
\small
 \centering 
 \renewcommand\tabcolsep{2pt}
 \renewcommand{\arraystretch}{1.2}	
\begin{tabular}{lccccccccccccc}
\toprule 
\multicolumn{2}{c}{}                                         &                                           & \multicolumn{11}{c}{\textbf{DataSets}}                                                                                                                                                                                                                                                                                                                                                                                                                           \\ \cline{4-14} 
\multicolumn{2}{c}{\multirow{-2}{*}{\textbf{Methods}}}       & \multirow{-2}{*}{\textbf{FLOPs ($\downarrow$)}}          & CIFAR-100                              & Pets                                   & CIFAR-10                               & Cars                                   & STL-10                                 & Fashion                                & RAF-DB                                 & Aircraft                               & Caltech                                & DTD                                    & Flowers                                \\ \hline\midrule
\multicolumn{2}{c}{\textit{\textbf{Linear-Probe:}}}          & \multicolumn{12}{c}{}                                                                                                                                                                                                                                                                                                                                                                                                                                                                                        \\
                              & Vanilla                      & 3.28\e{10}                                  & 57.55                                  & 89.95                                  & 77.57                                  & 52.85                                  & 93.98                                  & 87.82                                  & 52.12                                  & 20.97                                  & 83.93                                  & 62.61                                  & 87.24                                  \\
                              & MST                          & 3.28\e{10}                                  & 74.03                                  & 86.32                                  & 91.30                                  & 52.57                                  & 92.89                                  & 90.05                                  & 59.81                                  & 28.10                                  & 84.42                                  & 62.61                                  & 86.89                                  \\
\multirow{-3}{*}{ResNet50}    & \cellcolor{mygray}MSUN & \cellcolor{mygray}\textbf{3.12\ebf{10}} & \cellcolor{mygray}\textbf{74.97} & \cellcolor{mygray}\textbf{90.48} & \cellcolor{mygray}\textbf{92.40} & \cellcolor{mygray}\textbf{54.88} & \cellcolor{mygray}\textbf{94.86} & \cellcolor{mygray}\textbf{90.18} & \cellcolor{mygray}\textbf{60.37} & \cellcolor{mygray}\textbf{31.73} & \cellcolor{mygray}\textbf{84.52} & \cellcolor{mygray}\textbf{63.30} & \cellcolor{mygray}\textbf{88.37} \\
\hline 
                              & Vanilla                      & 2.28\e{10}                                  & 58.99                                  & 90.41                                  & 79.39                                  & 56.24                                  & 93.58                                  & 89.20                                  & 56.10                                  & 23.67                                  & 84.65                                  & 63.62                                  & 88.51                                  \\
                              & MST                          & 2.28\e{10}                                  & 74.22                                  & 90.14                                  & 91.67                                  & 56.40                        & 94.63                                    & 90.67                                  & 58.41                                  & 32.72                                  & \textbf{85.58}                         & 63.19                                  & 88.57                                  \\
\multirow{-3}{*}{\begin{tabular}[c]{@{}c@{}}DenseNet\\ 121\end{tabular}} & \cellcolor{mygray}MSUN & \cellcolor{mygray}\textbf{2.02\ebf{10}} & \cellcolor{mygray}\textbf{74.55} & \cellcolor{mygray}\textbf{90.82} & \cellcolor{mygray}\textbf{91.88} & \cellcolor{mygray}\textbf{58.22} & \cellcolor{mygray}\textbf{94.70} & \cellcolor{mygray}\textbf{90.76} & \cellcolor{mygray}\textbf{60.11} & \cellcolor{mygray}\textbf{33.54} & \cellcolor{mygray}85.42 & \cellcolor{mygray}\textbf{63.80} & \cellcolor{mygray}\textbf{89.88} \\
\hline 
                              & Vanilla                      & 1.24\e{11}                                  & 58.50                                  & 90.03                                  & 80.79                                  & 60.07                                  & 93.30                                  & 91.47                                  & 63.66                                  & 31.16                                  & 81.35                                  & 67.32                                  & 87.30                                  \\
                              & MST                          & 1.24\e{11}                                  & 74.65                                  & 89.35                                  & 91.85                                  & 56.04                                  & 95.42                                  & 92.82                                  & 64.37                                  & 34.33                         & 81.39                                   & 64.63                         & 84.89                                  \\
\multirow{-3}{*}{VGG16}       & \cellcolor{mygray}MSUN & \cellcolor{mygray}\textbf{1.13\ebf{11}}          & \cellcolor{mygray}\textbf{74.87} & \cellcolor{mygray}\textbf{90.24} & \cellcolor{mygray}\textbf{91.90} & \cellcolor{mygray}\textbf{60.75} & \cellcolor{mygray}\textbf{95.50} & \cellcolor{mygray}\textbf{93.33} & \cellcolor{mygray}\textbf{65.12} & \cellcolor{mygray}\textbf{34.45} & \cellcolor{mygray}\textbf{82.31} & \cellcolor{mygray}\textbf{68.11} & \cellcolor{mygray}\textbf{88.66} \\
\hline 
                              & Vanilla                      & 2.46\e{9}                                  & 51.35                                  & 88.56                                  & 74.99                                  & 51.34                         & 86.62                                    & 87.54                                  & 52.31                                  & 23.91                         & 83.86                                  & 61.86                                  & 87.54                                  \\
                              & MST                          & 2.46\e{9}                                  & 67.49                                  & 88.28                                  & 87.24                                  & 50.88                                  & 93.06                                  & 89.46                                  & 56.81                                  & 30.38                                  & 82.71                                  & 60.16                                  & 87.53                                  \\
\multirow{-3}{*}{\begin{tabular}[c]{@{}c@{}}MobileNet\\ V2\end{tabular}} & \cellcolor{mygray}MSUN & \cellcolor{mygray}\textbf{2.31\ebf{9}} & \cellcolor{mygray}\textbf{67.98} & \cellcolor{mygray}\textbf{89.82} & \cellcolor{mygray}\textbf{87.75} & \cellcolor{mygray}\textbf{51.63} & \cellcolor{mygray}\textbf{93.58} & \cellcolor{mygray}\textbf{89.72} & \cellcolor{mygray}\textbf{58.35} & \cellcolor{mygray}\textbf{31.32} & \cellcolor{mygray}\textbf{84.00} & \cellcolor{mygray}\textbf{62.32} & \cellcolor{mygray}\textbf{89.29} \\ \hline \midrule
\multicolumn{2}{c}{\textit{\textbf{Fine-tuned:}}}            & \multicolumn{12}{c}{}                                                                                                                                                                                                                                                                                                                                                                                                                                                                                        \\
                              & Vanilla                      & 3.28\e{10}                                  & 80.97                                  & 85.04                                  & 96.37                                  & \textbf{88.70}                         & 93.30                                  & 95.12                                  & 82.64                                  & \textbf{48.44}                         & 83.83                                  & 60.64                                  & 87.22                                  \\
                              & MST                          & 3.28\e{10}                                  & 82.63                                  & 84.39                                  & 96.61                                  & 83.77                                  & 93.70                                  & 95.43                                  & 81.23                                  & 42.98                                  & 88.94                                  & 60.68                                  & 88.42                                  \\
\multirow{-3}{*}{ResNet50}    & \cellcolor{mygray}MSUN & \cellcolor{mygray}\textbf{3.12\ebf{10}} & \cellcolor{mygray}\textbf{82.83} & \cellcolor{mygray}\textbf{86.42} & \cellcolor{mygray}\textbf{96.63} & \cellcolor{mygray}\textbf{88.39} & \cellcolor{mygray}\textbf{94.26} & \cellcolor{mygray}\textbf{95.51} & \cellcolor{mygray}\textbf{83.12} & \cellcolor{mygray}\textbf{49.31} & \cellcolor{mygray}\textbf{90.05} & \cellcolor{mygray}\textbf{61.29} & \cellcolor{mygray}\textbf{89.44} \\
\hline 
                              & Vanilla                      & 2.28\e{10}                                  & 81.46                                  & 85.61                                  & 96.23                                  & 89.00                                  & 93.07                                  & \textbf{95.24}                         & 80.80                                  & 49.94                                  & \textbf{85.54}                         & \textbf{60.35}                         & 87.37                                  \\
                              & MST                          & 2.28\e{10}                                  & 82.03                         & 86.43                                  & \textbf{96.98}                         & 87.35                                  & 94.76                                  & 95.65                                  & 82.53                                  & 48.92                                  & 89.44                                  & 60.00                                  & 90.31                                  \\
\multirow{-3}{*}{\begin{tabular}[c]{@{}c@{}}DenseNet\\ 121\end{tabular}} & \cellcolor{mygray}MSUN & \cellcolor{mygray}\textbf{2.02\ebf{10}} & \cellcolor{mygray}\textbf{82.75} & \cellcolor{mygray}\textbf{86.77} & \cellcolor{mygray}\textbf{96.81} & \cellcolor{mygray}\textbf{89.36} & \cellcolor{mygray}\textbf{94.83} & \cellcolor{mygray}\textbf{95.70} & \cellcolor{mygray}\textbf{83.04} & \cellcolor{mygray}\textbf{50.74} & \cellcolor{mygray}\textbf{89.94} & \cellcolor{mygray}\textbf{60.15} & \cellcolor{mygray}\textbf{90.63} \\
\hline 
                              & Vanilla                      & 1.24\e{11}                                  & 77.13                                  & 88.04                                  & 95.43                                  & 88.81                                  & 93.85                                  & 95.44                                  & \textbf{83.05}                         & 51.95                                  & 74.75                                  & 63.62                                  & \textbf{86.51}                         \\
                              & MST                          & 1.24\e{11}                                  & 79.84                                  & 88.26                                  & \textbf{96.25}                         & 87.16                                  & \textbf{94.95}                         & 95.54                                  & 82.63                                  & 51.38                                  & 82.44                                  & 62.55                                  & 89.82                                  \\
\multirow{-3}{*}{VGG16}       & \cellcolor{mygray}MSUN & \cellcolor{mygray}\textbf{1.13\ebf{11}} & \cellcolor{mygray}\textbf{80.97} & \cellcolor{mygray}\textbf{88.54} & \cellcolor{mygray}\textbf{96.70} & \cellcolor{mygray}\textbf{89.21} & \cellcolor{mygray}\textbf{94.87} & \cellcolor{mygray}\textbf{95.88} & \cellcolor{mygray}\textbf{82.69} & \cellcolor{mygray}\textbf{52.97} & \cellcolor{mygray}\textbf{83.15} & \cellcolor{mygray}\textbf{63.84} & \cellcolor{mygray}\textbf{89.98} \\
\hline 
                              & Vanilla                      & 2.46\e{9}                                  & 78.25                                  & 82.94                                  & 94.82                                  & 87.94                                  & 90.71                                  & 94.88                                  & 81.58                                  & 48.59                                  & 81.58                                  & 61.06                                  & 87.64                                  \\
                              & MST                          & 2.46\e{9}                                  & \textbf{80.93}                         & \textbf{83.95}                         & 95.67                                  & 86.05                                  & \textbf{92.90}                         & \textbf{94.99}                         & 81.03                                  & 48.23                                  & 87.72                                  & \textbf{61.81}                         & 89.14                                  \\
\multirow{-3}{*}{\begin{tabular}[c]{@{}c@{}}MobileNet\\ V2\end{tabular}} & \cellcolor{mygray}MSUN & \cellcolor{mygray}\textbf{2.31\ebf{9}} & \cellcolor{mygray}\textbf{80.49} & \cellcolor{mygray}\textbf{84.10} & \cellcolor{mygray}\textbf{95.95} & \cellcolor{mygray}\textbf{87.98} & \cellcolor{mygray}\textbf{93.24} & \cellcolor{mygray}\textbf{94.69} & \cellcolor{mygray}\textbf{82.10} & \cellcolor{mygray}\textbf{49.31} & \cellcolor{mygray}\textbf{88.11} & \cellcolor{mygray}\textbf{62.66} & \cellcolor{mygray}\textbf{89.76} \\ 
\bottomrule 
\end{tabular}
\end{table*}

\begin{table}[t]
\caption{The accuracy and FLOPs of each subnetwork across different input sizes. The model’s first, second, and third rows correspond to $f_1$, $f_2$, and $f_3$, respectively.}
 \label{table:3}
\small
 \centering 
 \renewcommand\tabcolsep{3pt}
 \renewcommand{\arraystretch}{1.2}	
\begin{tabular}{lcccccccc}
\toprule
\multirow{2}{*}{\textbf{Methods}} & \multirow{2}{*}{\textbf{FLOPs ($\downarrow$)}} & \multicolumn{7}{c}{\textbf{Input size}}                                                                       \\ \cline{3-9} 
                                  &                                 & 32            & 64            & 96            & 128           & 160           & 192           & 224           \\ \hline\midrule

\multirow{3}{*}{ResNet50}         & \textbf{2.84\ebf{10}}                        & \textbf{61.8} & 61.7          & 61.5          & 60.0          & 60.5          & 60.4          & 60.5          \\
                                  & 2.92\e{10}                        & 37.1          & \textbf{70.3} & \textbf{72.0} & \textbf{73.0} & 72.8          & 72.3          & 73.7          \\
                                  & 3.28\e{10}                        & 34.5          & 61.8          & 68.0          & 70.0          & \textbf{74.1} & \textbf{74.3} & \textbf{75.1} \\
\hline
\multirow{3}{*}{\begin{tabular}[c]{@{}c@{}}DenseNet\\ 121\end{tabular}}      & \textbf{1.60\ebf{10}}                        & \textbf{60.1} & 60.5          & 60.1          & 60.2          & 60.1          & 60.8          & 60.5          \\
                                  & 1.89\e{10}                        & 36.0          & \textbf{62.7} & \textbf{67.0} & \textbf{73.8} & 73.6          & 73.6          & 73.8          \\
                                  & 2.28\e{10}                        & 33.2          & 59.8          & 65.5          & 69.9          & \textbf{74.0} & \textbf{74.0} & \textbf{74.1} \\
\hline
\multirow{3}{*}{VGG16}            & \textbf{8.68\ebf{10}}                        & \textbf{58.9} & 58.5          & 58.9          & 57.9          & 58.9          & 58.1          & 58.9          \\
                                  & 9.83\e{10}                        & 28.4          & \textbf{59.0} & \textbf{68.0} & \textbf{72.8} & 72.5          & 72.4          & 72.8          \\
                                  & 1.24\e{11}                        & 27.6          & 53.4          & 65.0          & 68.7          & \textbf{72.9} & \textbf{73.0} & \textbf{74.0} \\
\hline
\multirow{3}{*}{\begin{tabular}[c]{@{}c@{}}MobileNet\\ V2\end{tabular}}      & \textbf{2.02\ebf{9}}                        & \textbf{58.1} & 57.8          & 58.1          
& 58.1          & 58.1          & 58.0          & 58.1          \\
                                  & 2.21\e{9}                        & 25.7          & \textbf{59.9} & \textbf{64.3} & \textbf{70.4} & 70.1          & 70.1          & 70.4          \\
                                  & 2.46\e{9}                        & 23.7          & 51.9          & 51.2          & 53.9          & \textbf{70.7} & \textbf{70.9} & \textbf{71.2} \\ 
\bottomrule
\end{tabular}
 \vskip -0.15in
\end{table}

\subsection{Comparative results}
This section discusses the results of experiments on multi-scale testing, linear-probe transfer, and fine-tuning transfer settings. 

\noindent\textbf{Multi-scale Testing:} 
In multi-scale testing (MST), the images in the dataset are rescaled into different sizes. All the rescaled images in training set are upsampled to the original size and used to train the model with fixed input size. When an image of smaller size is tested, it is upsampled to the standard input size of the model. Figure \ref{fig:4} and Table \ref{table:2} show the test accuracies on ImageNet dataset with different input sizes.
As results, MST surpasses Vanilla models at smaller input sizes, but sacrifices 4.03-5.20\% accuracy on larger sizes like \(224 \times 224\).
This is because the Vanilla model is trained with images of one size only, while in MST, the model is trained using images of multiple sizes.
Our method MSUN outperforms both Vanilla and MST across most input sizes, especially small sizes. For instance, in the case of ResNet-50, when employing MSUN, achieves an accuracy 61.83\% at the size of \(32 \times 32\) pixels, which is higher than MST’s 47.11\% and the Vanilla model’s 19.64\%. Similarly, DenseNet-121, VGG-16, and MobileNetV2 exhibit an accuracy boost of 8.92 - 17.78\% compared to MST, and 39.04 - 44.53\% when compared to Vanilla models. At the largest size \(224 \times 224\), the MSUN models experience only a slight drop of 0.12-0.73\% compared to the Vanilla models.
In summary, with a slight parameter increase of 1.56-11.43\%, MUSN achieves an 8.39-10.97\% rise in average accuracy and a 7.01-16.13\% cut in average FLOPs for input sizes ranging from \(32 \times 32\) to \(224 \times 224\).

\noindent\textbf{Linear-Probe Evaluation:} During linear evaluation, the feature extractor is fixed after training on ImageNet and linear classifiers are trained on target datasets.
As shown in Table \ref{table:4}, MST and MSUN exhibit significant improvements across the majority of models and datasets, indicating that enhancing multi-scale adaptability considerably boosts the transferability of models.
 Specifically, on lower image size datasets like CIFAR-10, CIFAR-100, Fashion-MNIST, and RAF-DB, MST and MSUN achieve 1.46-17.42\% and 0.71-16.48\% accuracy improvement respectively. 
For other datasets, MST boosts performance by 0.10-6.44\% in part of datasets and models but counters a 0.03-4.03\% drop in others.
MSUN exhibits robust performance, achieving 0.18-6.96\% improvements and declining 2.71\% only when using MobileNetV2 on the DTD dataset.
Moreover, it reduces average FLOPs by 3.55-8.45\% across these datasets.

\noindent\textbf{Fine-Tuning Transfer:} 
In this scenario, both the encoder and classifier of the model pre-trained on ImageNet are retrained together on the target dataset. 
As shown in Table \ref{table:4}, the disparity in accuracy between MST, MSUN, and standard models diminishes. Vanilla models exhibit a marginal advantage of 0.35-5.46\% on datasets such as DTD, RAF-DB, and Cars. 
MST improves accuracy by 0.10-7.69\% on the first six datasets, but reduces by 0.42-5.46\% on the remains. 
MSUN exhibits a 0.22-8.40\% improvement across all datasets, except for a minor 0.20-0.36\% drop on some models in the Cars and DTD dataset, demonstrating superior transferability. 
Additionally, it flexibly accommodates multi-scale inputs and reduces computational costs during retraining on the transfer datasets.

\begin{figure*}[h]
  \centering
   \includegraphics[width=1.0\linewidth]{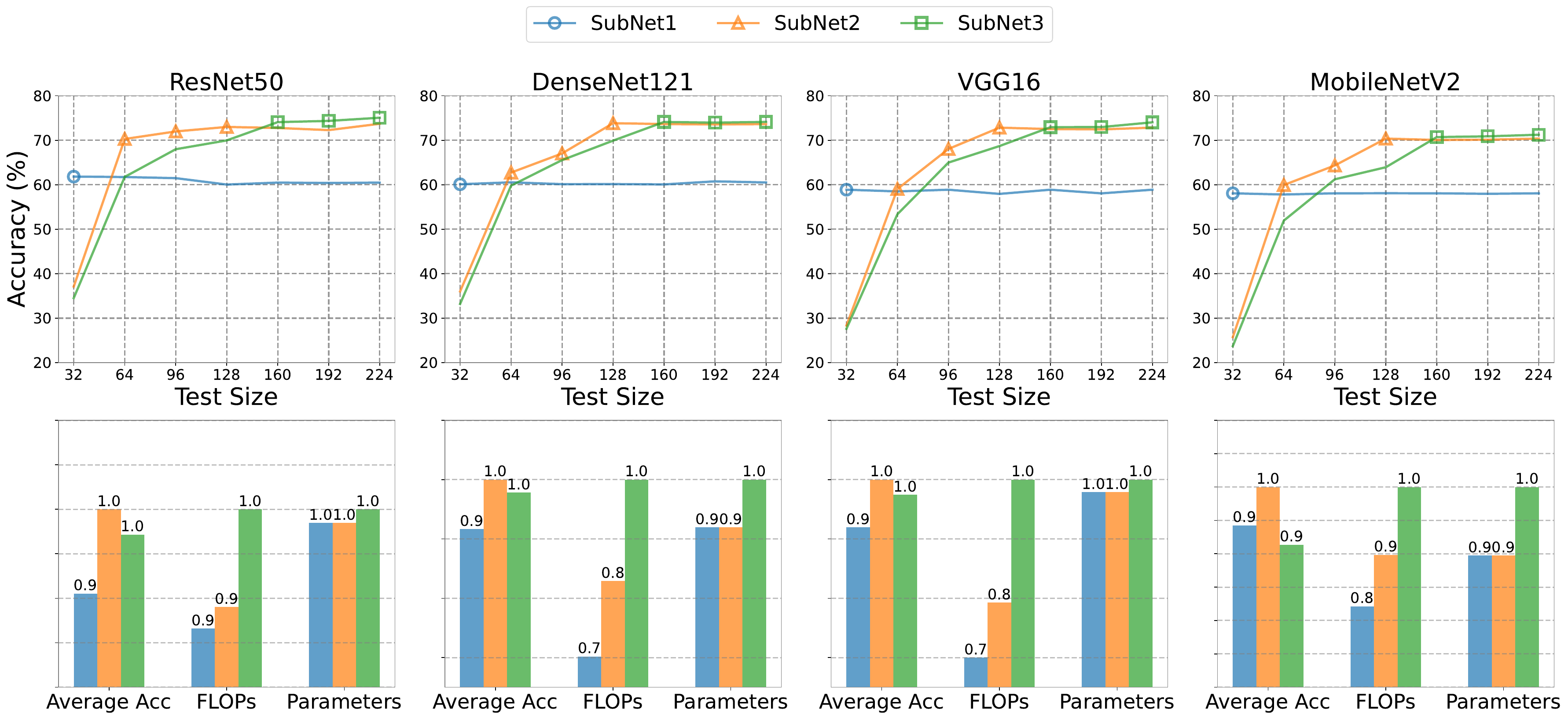}
  \vskip -0.1in
\caption{Comparison of multi-scale subnetworks. Tick marks denote optimal accuracy at each test size. (a) the accuracy curves with different input sizes.  (b) Mean accuracy, FLOPs, and total parameters across these inputs.}
  \vskip -0.13in
  \label{fig:5}
\end{figure*}

\subsection{Ablation study}

We analyze ablation studies utilizing the ImageNet dataset with ResNet-50 as the backbone model. The architecture of MSUN can be represented as $B$-blocks-$S$-subnets, where each sub-network is composed of a number of blocks (denoted as $B$), and the total number of sub-networks is denoted as $S$. For our implementation, we have employed one block and three sub-networks in our standard model, which is depicted as Res50-$B$1-$S$3. The ablation study consists of the following parts:

\noindent\textbf{Multi-scale Subnet Assessment:} We analyzed subnetworks $f_1$, $f_2$, and $f_3$, specialized for small, medium, and large-scale inputs respectively, to gauge their individual impact on performance. 
The results of multi-scale tests on ImageNet can be found in Table \ref{table:3} and Figure \ref{fig:5}.
$f_1$ achieves 24.1-34.4\% better performance for small scales input, such as \(32 \times 32\). 
Similarly, $f_2$ performs 1.5-8.5\% better in the mid-scale range of 64-128 and $f_3$ excels by 0.4-2.1\% for large scales between 160-224. 
The model adeptly handles varying input scales by utilizing multi-scale subnets derived from the network’s lower layers.
During inference, input is resized to the closest input scale of subnetworks, for extracting low-level features, as demonstrated by the green curve in Figure \ref{fig:5}.
Additionally, $f_1$, $f_2$ reduces FLOPs by 13.41-31.77\% and 10.16-19.18\% compared to a standard model ($f_3$), offering significant computational savings.

\begin{table}[t]
\caption{The accuracy of different test sizes under ImageNet multi-scale testing and model parameters, comparing distinct MSUN configurations.}
\label{table:5}
\small
\centering 
\renewcommand\tabcolsep{2.5pt}
\renewcommand{\arraystretch}{1.2}	
\begin{tabular}{lcccccccc}
\toprule
\multirow{2}{*}{\textbf{Methods}} & \multirow{2}{*}{\textbf{Params ($\downarrow$)}} & \multicolumn{7}{c}{\textbf{Input size}}        \\ \cline{3-9} 
                                  &                                  & 32   & 64   & 96   & 128  & 160  & 192  & 224  \\ \hline \midrule
Res50-B0-S3                       & 25.6M                            & 48.4 & 67.3 & 70.0 & 71.6 & 73.1 & 73.3 & 73.7 \\
Res50-B2-S3                       & 28.4M                            & 62.5 & 70.7 & 72.8 & 75.3 & 74.8 & 75.4 & 75.6 \\
Res50-B3-S3                       & 42.6M                            & 64.6 & 71.3 & 73.1 & 75.3 & 74.8 & 75.8 & 75.9 \\
Res50-B1-S2                       & 25.8M                            & 36.0 & 62.7 & 67.0 & 73.8 & 73.6 & 73.6 & 73.8 \\
Res50-B1-S4                       & 26.2M                            & 62.2 & 70.8 & 72.2 & 73.4 & 74.3 & 74.5 & 75.2 \\
Res50-B1-S5                       & 26.4M                            & 62.2 & 70.9 & 72.5 & 73.5 & 74.5 & 74.6 & 75.2 \\ \cdashline{1-9}
\rowcolor{mygray} 
Res50-B1-S3                       & 26.0M                            & 61.8 & 70.3 & 72.0 & 73.0 & 74.1 & 74.3 & 75.1 \\ 
\bottomrule
\end{tabular}
\vskip -0.15in
\end{table}

\begin{figure}[htb]
  \centering
  \vskip -0.1in
  \includegraphics[width=0.9\linewidth]{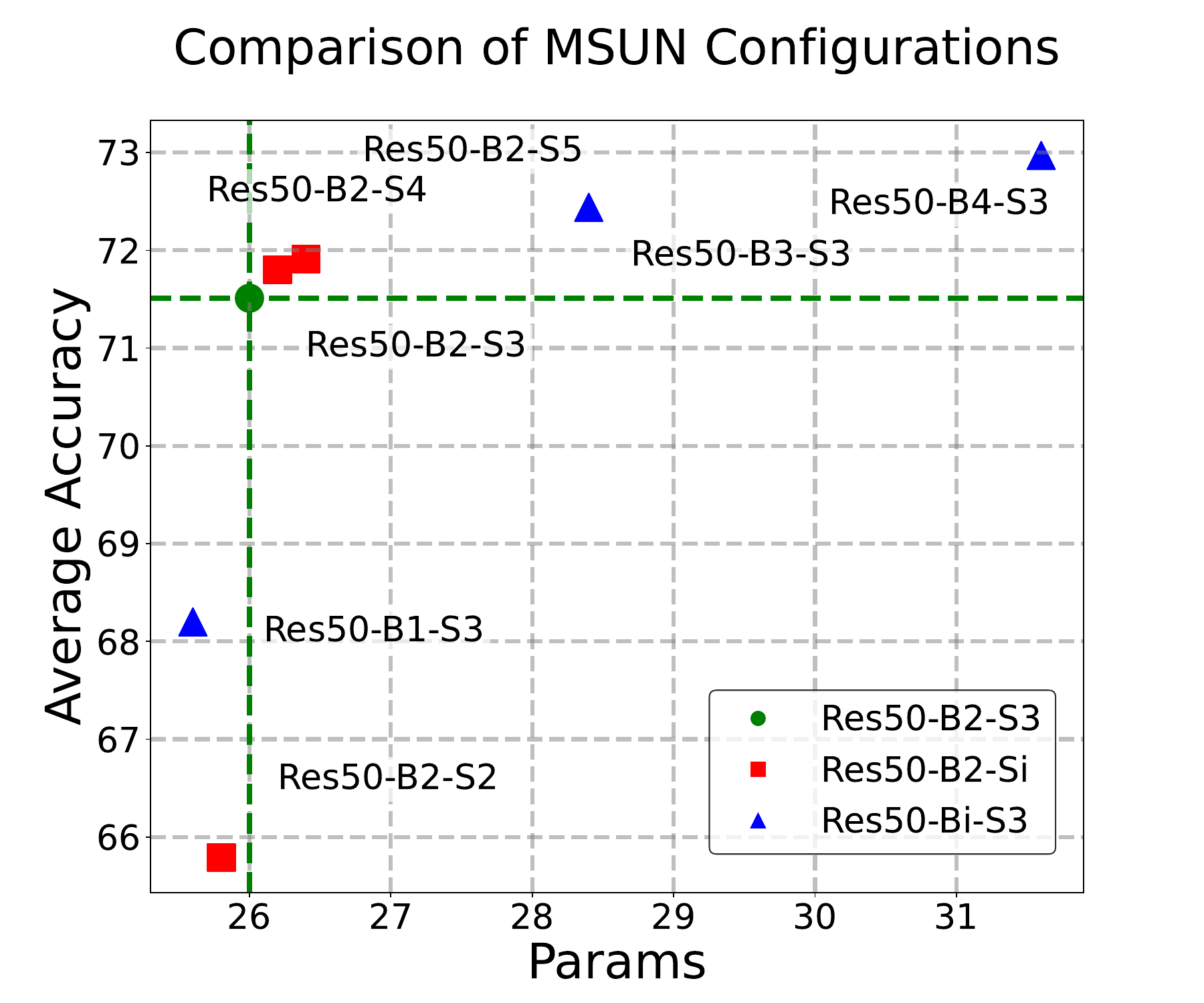}
  \vskip -0.1in
  \caption{The average accuracy and model parameters using different MSUN configurations. The red and purple lines indicate the parameters of the original model and the maximum achievable accuracy. Resnet50-B1-S3 maintains a good balance between parameters and accuracy.}
  \vskip -0.15in
  \label{fig:6}
\end{figure}

\noindent\textbf{Examining MSUN Configurations:} 
When employing our method, the CNN is decomposed into multi-scale subnetworks and a unified network. The model's configuration is defined by two primary factors: the number of blocks within each subnetwork and the total number of subnetworks employed. 
We tested ResNet50 in various settings, between 0 and 3 blocks and 2 to 5 subnetworks. The results of these tests are presented in Table \ref{table:5} and Figure \ref{fig:6}. 
It is shown that when increasing the number of blocks within a subnetwork, we observed that the MSUN approaches its upper bound, which is obtained by training an individual model for each image size.
However, increasing the number of blocks also leads to a corresponding parameter increase. Notably, with just a single block, the model achieved a performance increase of 19.7\%  with only a 1.56\%  increase in parameters. 
More blocks increase in parameters ranged from 10.94-66.41\%, resulting in performance improvements of 20.96-21.84\%. 
Likewise, Increasing the number of subnetworks also raises parameters and complexity. 
With two subnetworks, there is a 0.78\% increase in parameters while achieving 11.32-24.31\% performance gains on some input scales when compared to Vallina model. However, it results in a performance drop of 2.38\% on image size of \(224 \times 224\) (73.80\% versus 75.18\%). 
At block setting $B1$, increasing subnetworks from $S2$ to $S5$ improves the performance by 5.72-6.12\% and parameters by 0.77-2.32\%, and achieves accuracy comparable to the Vanillina model on \(224 \times 224\) size.

\begin{table}[t]
 \caption{The average accuracy on Imagenet multi-scale testing and model parameters, comparing use vs. not use multi-scale subnet, unified net, and scale-invariance constraint.}
 \label{table:6}
 \small
 \centering 
 \renewcommand\tabcolsep{3.5pt}
 \renewcommand{\arraystretch}{1.2}	
\begin{tabular}{lccccc}
\toprule
\textbf{Methods}       & \textbf{SI} & \textbf{SNet} & \textbf{UNet} & \textbf{Params ($\downarrow$)} & \textbf{Accuracy ($\uparrow$)} \\ \hline \midrule
MST             & \XSolidBrush  & \XSolidBrush    & \XSolidBrush   & 25.6M           & 66.76               \\
MST+SI              & \Checkmark  & \XSolidBrush    & \XSolidBrush   & 25.6M           & 66.95                \\
SNet            & \XSolidBrush& \Checkmark      & \XSolidBrush   & 76.7M           & 73.39              \\
SNet+SI         & \Checkmark  & \Checkmark      & \XSolidBrush   & 76.7M           & 72.84              \\
SNet+UNet       & \XSolidBrush& \Checkmark      & \Checkmark     & 26.0M           & 70.31              \\ \cdashline{1-6}\rowcolor{mygray} 
SNet+UNet+SI    & \Checkmark  & \Checkmark      & \Checkmark     & 26.0M           & 71.67              \\ 
\bottomrule
\end{tabular}
\end{table}

\noindent\textbf{Impact of Multi-scale Subnet, Unified Net, and SI:} 
We examined MSUN components, including the multi-scale subnetwork (SNet), unified network (UNet), and SI constraint, using the Res50-$B$1-$S$3 configuration. The average accuracy from \(32 \times 32\) to \(224 \times 224\) and parameter count are shown in Table \ref{table:6}.
It is evident that using both Snet and Unet results in slight parameter increase and large performance gain. Excluding Unet prevents sharing of networks across scales, resulting in 73.39\% accuracy but with a 199.61\% parameter increase. Combining Snet with Unet improves performance by 3.55\% with only a 1.56\% increase in parameters. Implementing the SI constraint increases Snet+Unet accuracy from 70.31\% to 71.67\%.
However, using the constraint alone (without SNet and UNet) only yields a marginal improvement of 0.19\% (MST versus MST+SI).
results indicate that the components SNet, UNet, and SI are crucial for improving the model's performance in multi-scale scenarios. Particularly, the UNet results in reduced parameter complexity while maintaining high performance.


\begin{figure*}[t]
  \centering
  \includegraphics[width=0.7\linewidth]{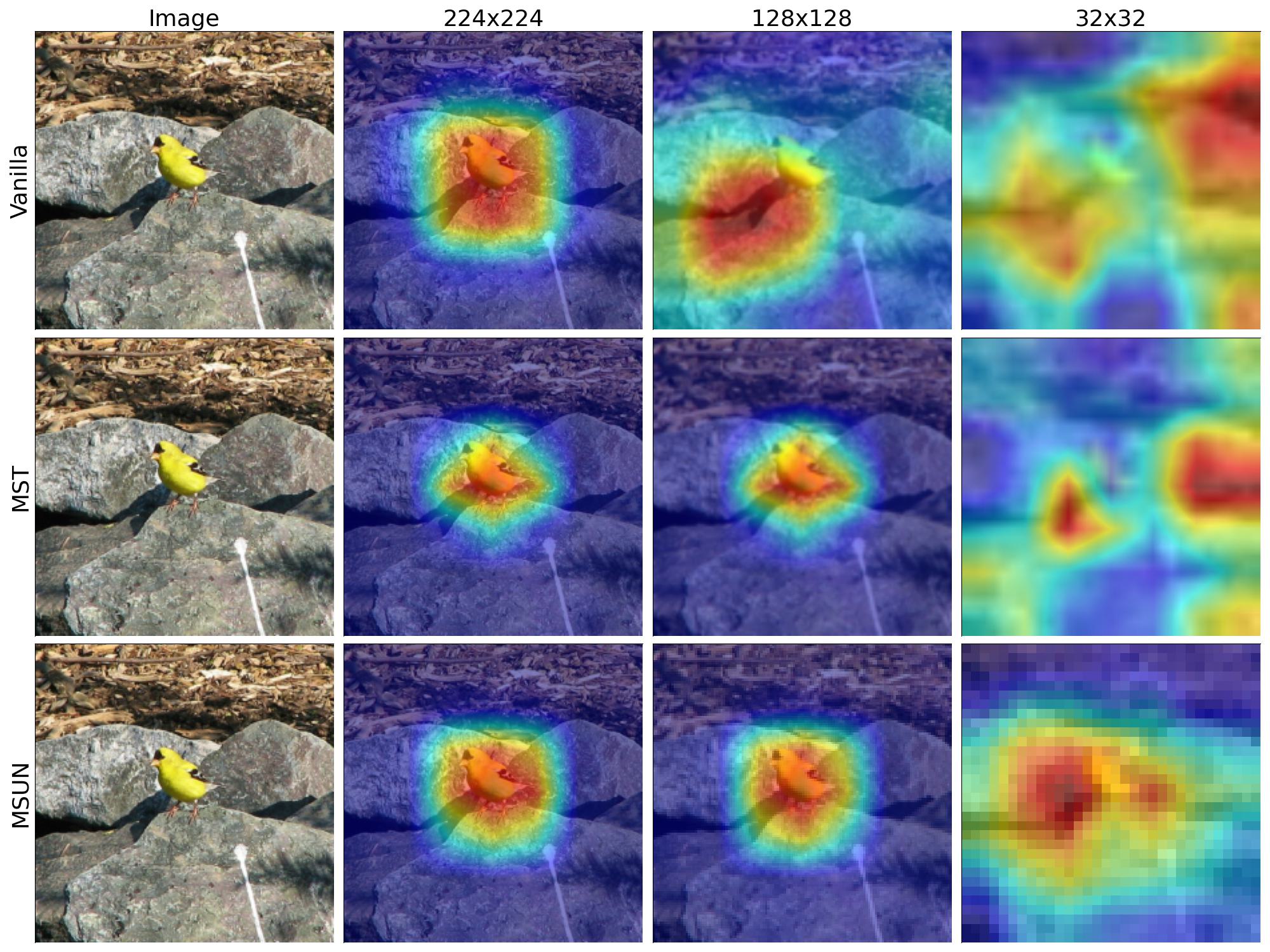}
  \vskip -0.1in
  \caption{Grad-CAM visualizations reveal that ResNet50 models pre-trained on ImageNet focus differently depending on input image size. The Vanilla model's attention region becomes inconsistent when the image size changes. MST's focus shifts relative to the Vanilla model, while MSUN retains consistent attention across different input sizes, aligning with the Vanilla model's focus on \(224 \times 224\) image size.}
  \vskip -0.1in
  \label{fig:7}
\end{figure*}

\begin{figure}[t]
  \centering
  \includegraphics[width=0.9\linewidth]{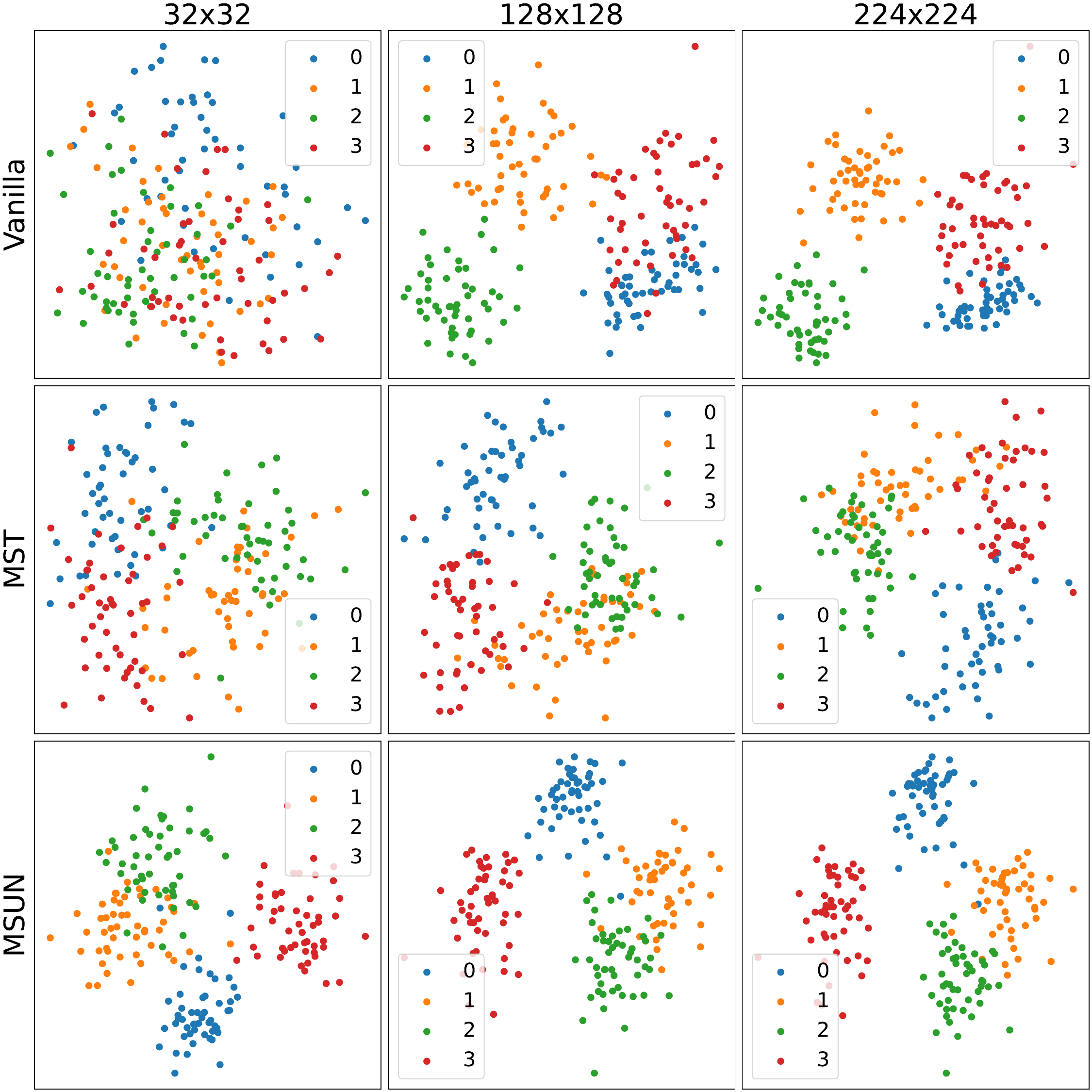}
  \vskip -0.1in
  \caption{PCA visualizations with a pre-trained ResNet50 on ImageNet. Vanilla confuses on \(32 \times 32\) image size, MST alters feature distribution, and MSUN maintains discriminative capabilities across different image size.}
  \vskip -0.15in
  \label{fig:8}
\end{figure}
\subsection{Further Analysis and Discussion}
\noindent\textbf{Visualization of attention area.}
Gaining insights into how models allocate attention to distinct areas within images is essential for assessing their robustness across varying image size. To this end, we employ Gradient-weighted Class Activation Mapping (Grad-CAM) as an analytical visualization tool to examine the model's focus on particular regions during decision-making. Grad-CAM is defined as follows:
\begin{equation}
L^c_{\text{Grad-CAM}} = \text{ReLU}\left(\sum_{k} \alpha_k^c A^k\right),
\end{equation}
where \( A^k \) is the features in the \( k \)-th channel of the last layer. \( \alpha_k^c \) is how much this feature affects class \( c \). \(\text{ReLU}\) keeps only positive features. As shown in Figure \ref{fig:7}, we use Grad-CAM on MSUN, Vanilla, and MST at \(224 \times 224\), \(128 \times 128\), and \(32 \times 32\) image size. The following phenomena are observed:

\begin{itemize}
  \item MSUN is highly robust to image size changes even on a lower resolution of \(32 \times 32\), where the focus areas largely overlap with those on the standard \(224 \times 224\) size with Vanilla.
  
  \item MST demonstrates consistency on \(128 \times 128\) image size but shows some defocus in the attention area on the smaller \(32 \times 32\) size. The attention region of the MST model differs largely from that of the Vanilla model.

  \item  The attention focus area of Vannila model appears highly sensitive to image size changes. While the model performs reliably at the \(224 \times 224\)  standard size, it loses focus consistency at lower size, indicating limited robustness.
\end{itemize}
In summary, MSUN shows high robustness over various image size, albeit with slight modifications in focus regions relative to the Vanilla model on large input size.

\noindent\textbf{Visualization of feature space.}
Through PCA visualization, we compared the feature spaces of the models at different input scales in Figure \ref{fig:8}. 
In the feature space of the Vanilla model on \(224 \times 224\) size, the features of different classes are well separated, but this discriminability degrades notably on \(32 \times 32\) size. 
MST mitigates this confusion on \(32 \times 32\) size, but also distort the feature landscape of the Vanilla model at higher image size, like \(224 \times 224\) and \(128 \times 128\), resulting in lower discrimination. Section 6.1’s experimental results show that this alteration impairs accuracy at elevated image size. 
In contrast, the MSUN model retains feature separation comparable to the Vanilla model at the \(224 \times 224\) image size and remains robust across multiple input scales.

\section{Conclusion}
\label{sec: 7}
To improve the robustness of features across variable input image scales, we propose Multi-Scale Unified Network (MSUN) —a restructuring of CNNs into multi-scale subnetworks within a unified network, guided by a scale-invariant constraint. MSUN ensures feature consistency, maintains high performance across different image scales, and significantly reduces computational overhead compared to multi-scale testing and training-test with a fixed large image size. The framework is readily applicable to various CNN architectures. We validated the effectiveness and superiority of the proposed method with a range of common CNN architectures on a number of datasets of various image sizes. Besides image classification evaluated in this paper, we expect that the proposed network can also apply to other computer vision tasks such as object detection, which deals with object regions of highly variable scales. In the future, we will also consider the multi-scale robustness design of other neural network architectures such as the transformer.

\bibliography{refer}
\bibliographystyle{plain}

\end{document}